\documentclass[sigconf]{acmart}
\AtBeginDocument{%
  \providecommand\BibTeX{{%
    \normalfont B\kern-0.5em{\scshape i\kern-0.25em b}\kern-0.8em\TeX}}}

\setcopyright{acmcopyright}
\copyrightyear{2018}
\acmYear{2018}
\acmDOI{10.1145/1122445.1122456}

\acmConference[Woodstock '18]{Woodstock '18: ACM Symposium on Neural
  Gaze Detection}{June 03--05, 2018}{Woodstock, NY}
\acmBooktitle{Woodstock '18: ACM Symposium on Neural Gaze Detection,
  June 03--05, 2018, Woodstock, NY}
\acmPrice{15.00}
\acmISBN{978-1-4503-XXXX-X/18/06}

\usepackage[ruled, lined, linesnumbered]{algorithm2e}
\SetKwInput{KwInput}{Input}
\SetKwInput{KwParams}{Parameters}
\usepackage{caption}
\usepackage{subcaption}
\begin{document}

\title{Label-informed Graph Structure Learning for Node Classification}

\author[Liping Wang, Fenyu Hu, Shu Wu, and Liang Wang]{Liping Wang$^{1,2,}$*, Fenyu Hu$^{1, 2}$*, Shu Wu$^{1,2,\dagger}$, and Liang Wang$^{1,2}$}

\makeatletter
\def\authornotetext#1{
 \g@addto@macro\@authornotes{%
 \stepcounter{footnote}\footnotetext{#1}}%
}
\makeatother

\authornotetext{The first two authors made equal contribution to this work.}
\authornotetext{To whom correspondence should be addressed.}

\affiliation{%
 \institution{\(^1\)Center for Research on Intelligent Perception and Computing, Institute of Automation, Chinese Academy of Sciences}
 \institution{\(^2\)School of Artificial Intelligence, University of Chinese Academy of Sciences}
 \country{}
}

\email{wangliping2019@ia.ac.cn,  fenyu.hu@cripac.ia.ac.cn}
\email{{shu.wu, wangliang}@nlpr.ia.ac.cn}

\def\authors{Liping Wang, Fenyu Hu, Shu Wu, and Liang Wang}

\begin{abstract}
Graph Neural Networks (GNNs) have achieved great success among various domains. 
Nevertheless, most GNN methods are sensitive to the quality of graph structures. 
To tackle this problem, some studies exploit different graph structure learning strategies to refine the original graph structure. 
However, these methods only consider feature information while ignoring available label information. 
In this paper, we propose a novel label-informed graph structure learning framework which 
incorporates label information explicitly through a class transition matrix. 
We conduct extensive experiments on seven node classification benchmark datasets
and the results show that our method outperforms  or matches the state-of-the-art baselines.
\end{abstract}





\maketitle

\section{Introduction}
As a powerful tool of analyzing graph-structured data, Graph Neural Networks (GNNs) have recently demonstrated great success across various domains, including node classification \cite{gcn}, link prediction \cite{gnn_linkprediction}, 
recommendation systems \cite{srgnn}, etc. 
Despite GNNs' powerful ability in learning expressive node embeddings,
these methods are sensitive to the quality of graph structures.
To be more specific, graphs in the real world are often noisy due to the error-prone data-collection process.  
For example, in a citation network, a paper may include citations to irrelevant papers or miss citations to highly relevant papers.  
Since GNNs recursively aggregate neighborhood information across edges to obtain node embeddings, the above noise in the graph will propagate to a lot of neighborhood nodes, hindering the performance. 

Recently, some studies \cite{idgl, lds} attempt 
to boost the performance 
of GNNs through jointly learning a denoised graph structure and node embeddings.
These works can be unified under Graph Structure Learning (GSL) \cite{zhu2021deep}.
The key rationale behind these works is 
to remove the suspicious or add a potential edge between two nodes according to the distance or similarity between their embeddings. 
For example, IDGL \cite{idgl} first computes weighted cosine similarity between node embeddings. 
Then, this similarity is used to refine the original graph structure. 
Lastly, the optimal graph structure can be acquired by  directly optimizing downstream tasks such as node classification or link prediction. 


However, all of the existing GSL methods ignore available label information during the graph structure learning process.
A potential edge between two nodes is added to the graph if they have similar features or embeddings regardless of their labels. 
These added edges may contain noise and be harmful to the performance.
Take a citation network as an example, two papers focusing on the same problem adopt totally different approaches, thus they should be classified into two different categories. 
Since these two papers co-cite some classic papers solving the same problem, they have some common neighbors in the citation network.
Accordingly, the distance between their embeddings learned by GNNs is relatively short.
In this case, existing GSL methods tend to add an edge between them,
misleading the model to classify them into the same category.


To overcome this limitation, we propose a label-informed graph structure learning framework (LGS) which \textit{incorporates label information into graph structure learning explicitly}. 
To be more specific, 
we employ a class transition matrix, where each element represents 
the probability of an edge between nodes of two classes. 
Different from existing GSL methods, 
we consider feature similarity and class transition probability at the same time.
Intuitively,  for two nodes with very similar features, 
if the transition probability between their corresponding classes is very low, it is still not appropriate to add an edge between them.
In contrast, if  the transition probability is very high, there may still be an edge between them even if their features are dissimilar. 
Still using the above citation network to illustrate, since the two papers adopt totally different approaches, 
there are usually no direct citations between them.
Considering their label information, it is less likely for LGS to add an edge between them due to much lower transition probability.
In LGS, label information serves as an informative supplement to feature similarity. 

The main contributions of this work are summarized as follows:
\begin{itemize}
    \item Apart from feature similarity, we explicitly consider label information in graph structure learning. We introduce a novel iterative graph structure learning framework for node classification.
    \item We conduct extensive experiments on both homophily and heterophily graph datasets, demonstrating the superiority of our method.
\end{itemize}

\section{Methodology}

\subsection{Prelinimary}
\textbf{Problem Formulation.} Given a graph with an adjacency matrix $A \in \mathbb R^{n\times n}$ and a feature matrix $X\in \mathbb R^{n\times d}$.
$V_l$ is the set of labeled nodes.
Since the original graph structure may be noisy and incomplete, the goal is to learn the optimal graph structure and make predictions for unlabeled nodes simultaneously.
\subsection{Overview of LGS Framwork}

\begin{figure}
\centering
\includegraphics[width=\linewidth]{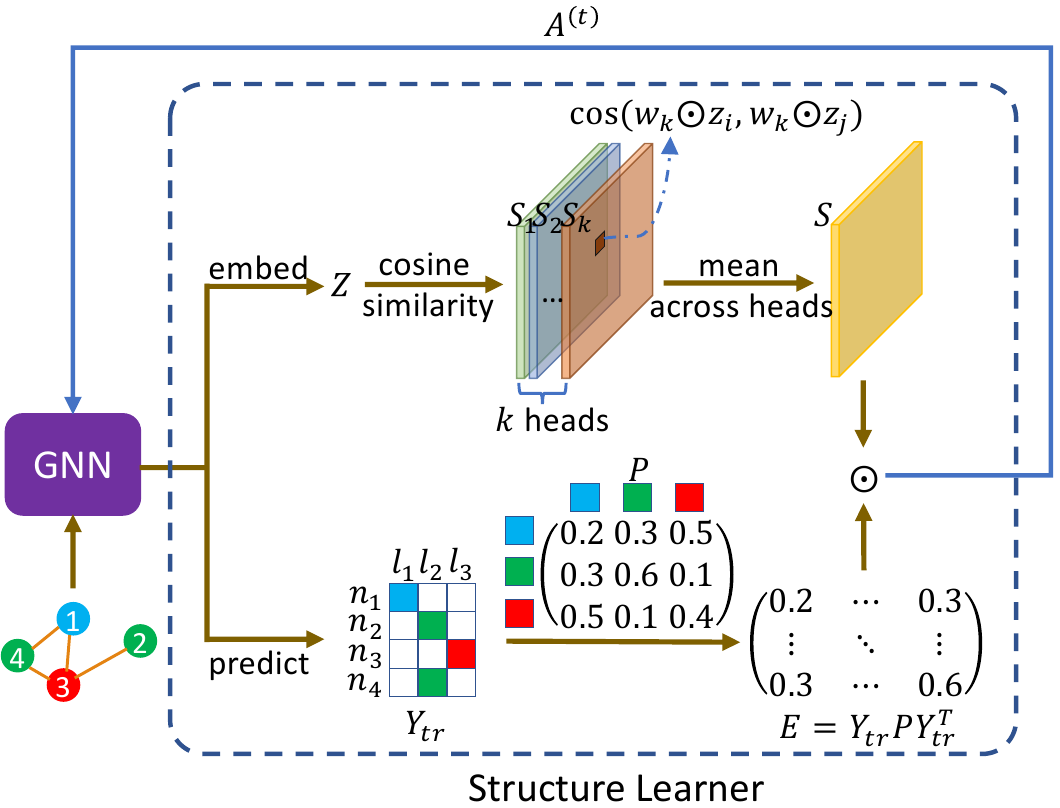}
\caption{Framework of LGS.}
\label{fig:sketch}
\end{figure}

As illustrated in Figure \ref{fig:sketch}, LGS consists of a GNN and a structure learner.
The GNN acts as a feature extractor and a classifier at the same time.
On the one hand,  the GNN outputs intermediate results $Z$ generated by its last hidden layer as node embeddings which encode feature information.
On the other hand, the GNN makes predictions for unlabeled nodes. 
Combining with the ground truth of labeled nodes, the GNN generates (pseudo) labels $Y_{tr}$ for all the nodes. 
There are two branches in the graph structure learner, which consider feature information and label information respectively.
The first branch computes multi-head weighted-cosine similarity between  each pair of nodes according to their embeddings.
Then, a feature similarity matrix is obtained by computing the mean across multiple heads.
The second branch generates an edge probability matrix $E\in \mathbb R^{n\times n}$ based on  (pseudo) labels $Y_{tr}$ and a class transition parameter matrix $P\in \mathbb R^{c\times c}$.


\subsection{GNN Architecture}

Without loss of generality, we choose two representative GNN architectures as feature extractor: GCN \cite{gcn} and ChebNet \cite{chebnet}.
For GCN, the graph convolution in the $l$-th layer can be described as:
\begin{align}
    \begin{split}
         H^{(l)} &=\sigma\left(\tilde{A} H^{(l-1)} W^{(l-1)}\right),\\
    \tilde{A} &=\hat{D}^{-\frac{1}{2}} \hat{A} \hat{D}^{-\frac{1}{2}},
    \end{split}
\end{align}
where $\hat A = A + I$ is the adjacency matrix of graph with self-loops, $\hat D$ is its corresponding degree matrix with $\hat D_{ii}=\sum_j \hat A_{ij}$, and $\sigma$ is non-linear activation function such as ReLU. 
As to ChebNet, the computation can be formulated as: 
\begin{align}
    \begin{split}
         H^{(l)} &= \sigma\left(\sum_{k=0}^{K} \theta_{k} T_{k}(\tilde{L}) H^{(l-1)}\right),\\
         \tilde{L} &= \frac{2 L}{\lambda_{\max }}-I,
    \end{split}
\end{align}
where $L= D^{-\frac{1}{2}}(D-A)D^{-\frac{1}{2}}$ is graph Laplacian matrix and $T_{k}$ is the $k$-th order Chebshev polonomial. 

In order to incorporate class transition matrix $P$ into graph convolution explicitly,
we add a label propagation layer weighted with $P$ at the end of GNN similar to \cite{cpgnn}.
In conclusion, the output of GNN is formulated as:
\begin{equation}
	(Z, \hat Y) = GNN(A, X),
\end{equation}
in which $A$ is an adjacency matrix, and $X$ is a feature matrix.

\subsection{Label-informed Graph Structure Learning}
\textbf{Feature Similarity Matrix}. Although there are various options 
for distance or similarity computation, 
such as Euclidean distance, attention mechanism, Mahalanobis distance and cosine similarity.
Without loss of generality, we adopt weighted consine similarity as metric function.
To further enrich expressiveness, we adopt a multi-head manner similar to GAT\cite{gat}. Specifically, in the $k$-th head,  the $n\times n$ similarity matrix $S_{k}$ is given by: 
\begin{equation}\label{eq:sim_1}
    S_{k}[i][j] = \cos(w_k\odot z_i, w_k\odot z_j),
\end{equation}
where $\odot$ is element-wise product operator, $w_k$ is a trainable weight and $z_i, z_j$ are the $i$-th  and $j$-th rows of $Z$, representing embeddings for node $v_i, v_j$ respectively. 
Then, a feature similarity matrix $S\in \mathbb R^{n\times n}$is obtained by:
\begin{equation}\label{eq:sim_2}
    S = \frac{1}{K} \sum_{k=1}^K S_k.
\end{equation}

\textbf{Class Transition Matrix.} In feature similarity matrix, only feature information of nodes is considered, while label information is ignored.  
To make full use of available label information, 
we employ a trainable matrix $P \in \mathbb R^{c\times c}$ to reweight similarity score between nodes, where $c$ is the number of classes of node. 
Intuitively, $P_{s,t}$ could be interpreted as 
the probability that an edge exists between  
a $s$-th class node and a $t$-th class node.

In order to reduce the difficulty of optimization, we investigate the initialization strategy for class transition matrix $P$.
According to the definition of $P$, $Y_{tr}^T A Y_{tr}$ serves as a good un-normed estimation.
Considering that $P$ should satisfy the double stochastic property (each row and each column sums to one), 
we propose to adopt the Sinkhorn-Knopp\cite{sinkhorn1967concerning} algorithm which operates iteratively to generate a double stochastic matrix.
So class transition matrix $P$ is initialized as $\text{Sinkhorn-Knopp}(Y_{tr}^T A Y_{tr})$.

\textbf{Learning Graph Structure}. 
Given a class transition matrix $P$, the probability of edges between each pair of nodes can be obtained according to their labels.
Nevertheless, labels of most nodes  are unavailable, so we assign pseudo labels to them according to  predictions of the current model. 
Formally, let $Y\in \{0, 1\}^{n\times c}$ be ground truth matrix where each row is an one-hot vector responding to each node, and $M\in \{0, 1\}^{n}$ be mask for labeled nodes. 
Then we define $Y_{tr}$ as:
\begin{equation}
\label{eq:y_tr}
    Y_{tr} = M\odot Y + (1-M) \odot \hat Y.
\end{equation}
The $i$-th and $j$-th row of $Y_{tr}$ represent (pseudo) labels of nodes $v_i, v_j$ respectively.
Only considering label information, the probability of an edge between nodes $v_i, v_j$ is $y_i^TPy_j$ according to the random walk theory\cite{klenke2013probability}.
Using matrix notation, this can be formulated as:
\begin{equation}
\label{eq:label}
	  E = Y_{tr} P Y_{tr}^T.
\end{equation}

The simplest way to combine feature similarity matrix $S$ and edge probability matrix $E$ in graph structure learning is through element-wise product $S\odot E$.
However, empirically we find this would make training unstable.
Hence, we introduce a hyper-parameter $r$ to control the weight of $E$.
In the real world, underlying graph structures  are relatively sparse than fully-connected graphs 
which not only include noise, but also are computationally expensive.
In addition, elements of a typical adjacency matrix are non-negative. Hence, we obtain a sparse non-negative matrix through $\epsilon$-neighborhood sparsification, which masks elements less than $\epsilon$(a non-negative hyper-parameter controlling sparsity) to zero. 
In summary, the refined adjacency matrix can be formulated as:
\begin{equation}
\label{eq:sparse}
      \widetilde A =\epsilon\text{-neighborhood}(S \odot (r * E + (1-r) *  \mathbf{1})),
\end{equation}
where $\mathbf{1}$ is the all ones matrix with the same shape to $E$.

\begin{algorithm}
\SetAlgoLined
\KwInput{$X, A, M, Y$}
\KwParams{$\alpha, \beta, \epsilon, r$}
\KwResult{$\hat Y^{(t)}$}
$\hat Y^{(0)}, Z^{(0)}\gets $ \textbf{GNN}(A, X)\\
compute $S_f$ according to $X$ following Eq \ref{eq:sim_1}, \ref{eq:sim_2} \\
 \For{$t\gets 1$  \KwTo $T$}{
compute $S$ according to $Z^{(t-1)}$ following Eq \ref{eq:sim_1}, \ref{eq:sim_2} \\
 $Y_{tr} \gets M\odot Y + (1-M) \odot \hat Y^{(t-1)}$ \\
 $E \gets Y_{tr}PY_{tr}^T$\\
$ A^{(t)} \gets \epsilon\text{-neighborhood}(S\odot (r*E + (1-r)*\mathbf{1}))$\\
$\widetilde A^{(t)} \gets \alpha A  + \beta  S_f + (1-\alpha - \beta) A^{(t)}$\\
$\hat Y^{(t)},  Z^{(t)} \gets \textbf{GNN}(\widetilde A^{(t)}, X)$\\
 }
$\mathcal L \gets  L_c(\hat Y^{(0)}, Y) + \frac{1}{T}\sum_{t=1}^TL_c(\hat Y^{(t)}, Y) + \Phi(P)$\\
back-propagating through $\mathcal L$ to update parameters
\caption{Training of LGS}
\end{algorithm}

\subsection{Training}

\textbf{GNN Warm-up}. In order to obtain relatively accurate pseudo label for unlabeled nodes, the GNN is trained solely for several epochs with classification loss function
\begin{equation}
    L_c(\hat Y, Y) = \sum_{i\in  V_l} \text{CE}(\hat Y_i, Y_i), 
\end{equation}
where $\hat Y_i$ is the prediction of the GNN for node $v_i$, and CE denotes cross entropy loss.

\textbf{Iterative Graph Structure Learning}. The graph structure learner and  the GNN are jointly optimized in an iterative manner for $T$ steps. 
In the $t$-th iteration, given node embedding $Z^{(t-1)}$ and predictions $\hat Y ^{(t-1)}$computed in the $(t-1)$-th iteration, the graph structure learner compute refined adjacency matrix $A^{(t)}$.
Although the original graph structure may be inaccurate and incomplete, it still carries relatively rich and useful information. 
What's more, empirically, we find that feature similarity matrix $S_f$ computed according to raw feature $X$ serves as a relatively accurate refinement to the original graph structure. 
As a result, we combine $A$,  $S_f$ and $A^{(t)}$ together:
\begin{equation}
\label{eq:t_step}
     \widetilde A^{(t)} = \alpha A  + \beta  S_f + (1-\alpha - \beta) A^{(t)},
\end{equation}
in which $\alpha$ and $\beta$ are hyper-parameters that control relative importance assigned.
Based on the refined graph structure $\widetilde A^{(t)}$, the GNN outputs node embeddings $Z^{(t)}$ and predictions $\hat Y^{(t)}$ for next iteration use.
 
\textbf{Joint Optimization.}
After $T$ iterations, total loss function is given by:
\begin{align}
\begin{split} 
\Phi(P) = &\sum\nolimits_i \left | \sum\nolimits_j P_{ij}\right |,\\
\mathcal L = L_c(\hat Y^{(0)}, Y) + &\frac{1}{T}\sum_{t=1}^TL_c(\hat Y^{(t)}, Y) + \Phi(P), 
\end{split}
\end{align}
in which $L_c$ is cross-entropy classification loss and $\Phi(P)$ is a regularization item to encourage the sum of each row of transition matrix $P$ to center around zero.
Then, the GNN and the graph structure learner are optimized through common gradient descent algorithms.

\section{Experiment}
In this section, we conduct extensive experiments to verify the effectiveness of the proposed method LGS for node classification on both homophily and heterophily \cite{geomgcn} graph datasets.

\subsection{Setup}
\begin{table}
\centering
\caption{Data Statistics}
\label{tbl:data}
\resizebox{\linewidth}{!}{
\begin{tabular}{lccccccc}
\toprule
    & Cora   & Citeseer       & Cornell    & Chameleon  & Squirrel & Wisconsin   & Texas  \\
\midrule
Homophily Ratio  & 0.81  & 0.74   & 0.3     & 0.23   & 0.22  & 0.21   & 0.11  \\
\# Nodes    & 2,708 & 3,327  & 183      & 2,277  & 5,201  & 251    & 183 \\
\# Edges    & 5,278   & 4,676  & 280   & 31,421  & 198,493   & 466   & 295    \\
\# Features & 1,433   & 3,703    & 1,703     & 2,325  & 2,089 & 1,703  & 1,703    \\
\# Classes  & 7 & 6   & 5     & 5  & 5  & 5 & 5 \\
\bottomrule
\end{tabular}
}
\end{table}

\begin{table*}[]
\centering
\caption{Node classification accuracies.}
\label{tbl:result}
\begin{tabular}{lccccccc}
\toprule
    & Cora                                                          & Citeseer                               & Cornell                                & Chameleon                              & Squirrel                                            & Wisconsin                                           & Texas                                               \\
\midrule
GCN         & 86.66 $\pm$ 1.45                      & 76.25 $\pm$ 1.19  & 59.73   $\pm$ 6.33 &  38.99 $\pm$ 1.86 &  29.20 $\pm$   1.10 &  53.92 $\pm$   5.49 & 59.73 $\pm$   6.33 \\
ChebNet     & $86.14 \pm 1.35$                                              &  $76.34 \pm   1.59$ & $74.86 \pm 8.02$                       &  $46.05   \pm 1.46$ &  $30.30 \pm   1.50$ & $76.08 \pm 2.29 $                      & $74.59 \pm 8.04$                       \\
GAT         &  $87.20 \pm   1.09$                       &  $75.92 \pm   1.59$ &  $59.46   \pm 4.01$ &  $44.06   \pm 2.52$ & $ 27.49 \pm   1.52$ &  $54.71 \pm 3.66$  &  $58.65 \pm   4.84$ \\
\hline
GEOM-GCN & $85.26 \pm 1.57$ & \textbf{77.99$\pm$1.25} & $60.54\pm 3.67$ & $60.00\pm 2.81$ & $38.15\pm 0.92$ & $64.51\pm 3.66$ & $66.76\pm 2.72$\\
CPGNN       &   $87.00 \pm   1.02$ &  $76.07 \pm   1.21$ &  \underline{$75.14   \pm 7.43$} &  \underline{$62.21   \pm 3.29$} & \underline{$ 40.16 \pm   6.43$} &  \underline{$76.47 \pm   2.77$} & \underline{$ 75.68 \pm   7.15$} \\
\hline
IDGL        & \underline{$87.28 \pm 1.00$}                                         &  $76.88 \pm   1.64$ &  $68.11   \pm 8.87$ &  $38.51   \pm 4.65$ &  $25.18 \pm   2.40$ &  $55.69 \pm   4.57$ &  $66.49 \pm   6.07$\\
Pro-GNN     & $83.52 \pm 2.20$   & $72.96 \pm 1.99$                             & $62.97\pm 7.93$                        & $58.25\pm 3.84$                        & $32.59\pm 1.04$                        & $55.69\pm 5.96$                        & $60.81\pm 6.07$                        \\
\hline
LGS-GCN    & \textbf{87.38 $\pm$ 1.25}                                              & \underline{76.92 $\pm$ 1.75}                       & $63.78 \pm 7.76$                       & $56.27\pm 3.16$                        & $34.92\pm 2.21$                        & $53.14\pm 6.70 $                   & $59.73 \pm 6.99$                       \\
LGS-Cheb   & 86.14 $\pm$1.35                                         & 76.65 $\pm$ 1.78   & \textbf{76.76 $\pm$ 8.56}   & \textbf{71.45 $\pm$ 2.17}  & \textbf{48.94 $\pm$ 4.44} & \textbf{76.86 $\pm$ 3.70}  & \textbf{75.95 $\pm$ 7.69}\\
\bottomrule
\end{tabular}
\end{table*}

\textbf{Datasets}. For homophily graphs, we choose two citation networks, Cora and Citeseer \cite{sen2008collective}. 
For heterophily graphs, we choose Chameleon, Squirrel, Wisconsin and Texas \cite{geomgcn}.
Statistics for these datasets could be found  in Table \ref{tbl:data}, 
where  the homophily ratio of a graph represents the tendency of a node to have nodes of the same class as its neighbors, and can be computed as:
\begin{equation}
 h_G =\frac{1}{n}\sum_{i=1}^n h_i  = \frac{1}{n}\sum_{i=1}^n\frac{|N_i^s|}{|N_i|},
\end{equation}
where $h_i$ represents homophily ratio of node $v_i$ and $N_i^s$ is the set of $v_i$' neighboring nodes with the same label to $v_i$.
Low homophily corresponds to high heterophily.
For all datasets, we follow the data splits given in Geom-GCN \cite{geomgcn}.

\textbf{Baselines}. 
We compare our methods with following methods from three categories: (1) classic GNN models for node classification: GCN \cite{gcn}, ChebNet \cite{chebnet} and GAT \cite{gat}, (2) recent methods designed specially for heterophily graphs: GEOM-GCN \cite{geomgcn} and CPGNN \cite{cpgnn},
and (3) the state-of-the-art models with graph structure learning: IDGL \cite{idgl} and Pro-GNN \cite{pro-gnn}.
\subsection{Implementation}

Even though our framework is agnostic to the choice of specific GNN architecture, we choose two representative GNNs: GCN and ChebNet, and the corresponding model variants are termed as LGS-GCN and LGS-Cheb respectively. 

For a fair comparison, we implement our method and all baselines in the same experimental settings as \citet{geomgcn}.
 We run all methods on all ten splits, and report mean and standard deviation of accuracies on the test set.
For methods with multiple variants like CPGNN \cite{cpgnn}, the best performance is reported. 

For hyper-parameter setting, we set the embedding dimension to 64, the number of layers to 2, $\epsilon$ to zero. 
And $\alpha$ is fixed at $0.8$. We train the model using Adam optimizer \cite{adam14} with an initial learning rate of 0.01. 
Moreover, for all the datasets, we first train the GNN alone for 400 epochs, then train the GNN and the graph structure learner for 1600 epochs together. 


\subsection{Main Results}

Mean and standard deviation of accuracies for node classification on test sets over 10 splits are reported in Table \ref{tbl:result}. 
Our method obtains best performance on almost all the datasets with varing homophily ratios.
Compared with graph structure learning (GSL) methods considering only feature information, our method outperforms them by a wide margin, reflecting the necessity to take available label information into consideration. 

Compared with GEOM-GCN and CPGNN designed specially for graphs with strong heterophily, our method still achieves significant improvement, owing to the refined graph structure by considering both feature information and label informaiton. 

Notably, ChebNet outperforms GCN by a wide margin on graphs with high hetetrophily ratios, and is slightly inferior on homophily graphs. 
As analyzed in \cite{shuman2013emerging}, GCN implicitly treats high-frequency components as ``noises'', and has them discarded. 
However, this may hinder the generalizability since high-frequency components can carry meaningful information about local discontinuities, 
This could also explain why LGS-Cheb performs better than LGS-GCN on heterophily graphs like Chameleon, Squirrel, etc. 

\begin{figure}
    \centering
\includegraphics[width=\linewidth]{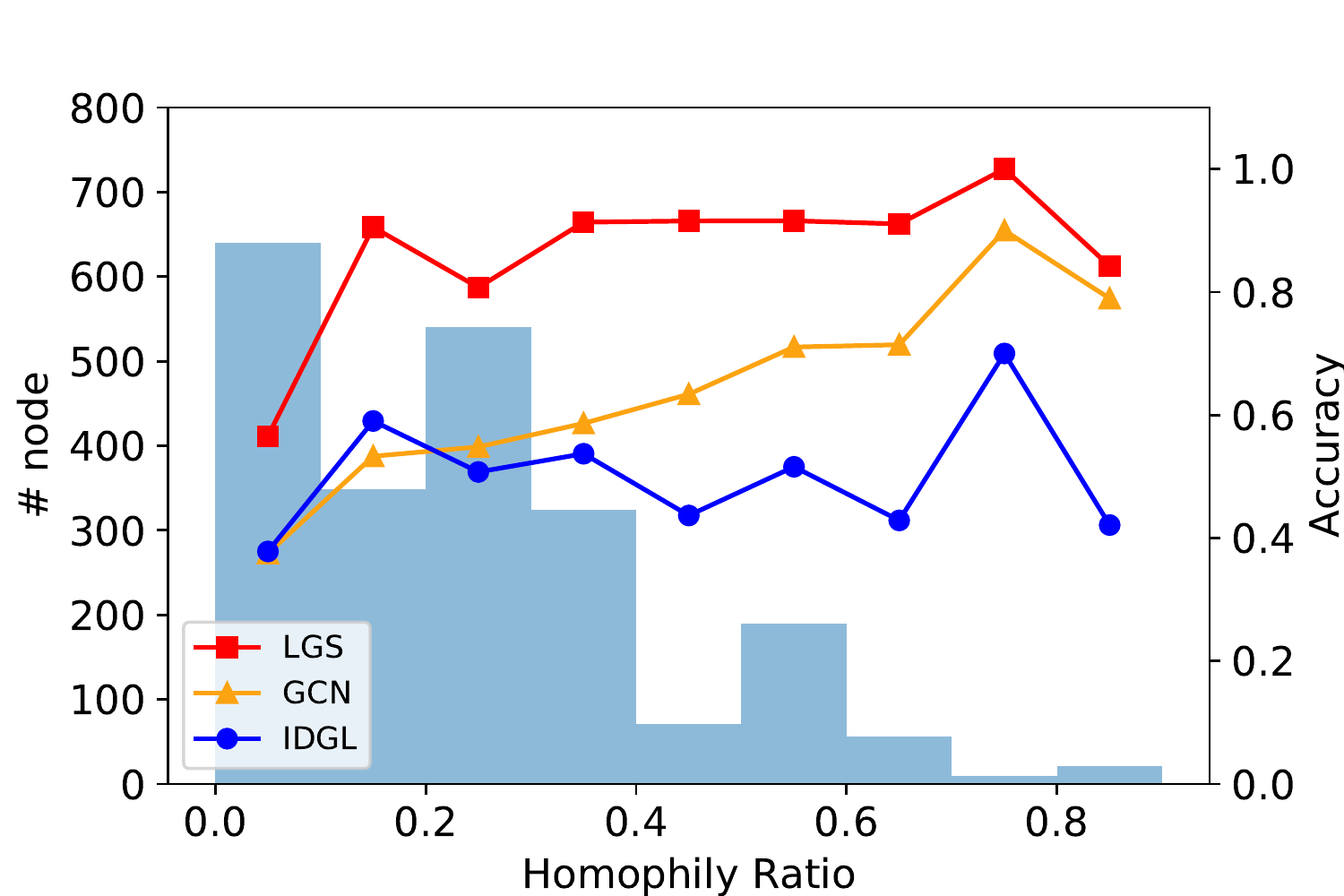}
    \caption{Distribution of nodes with homophily ratio and classification accuracy for LGS, GCN and IDGL on Chameleon dataset.}
    \label{fig:acc}
\end{figure}

\subsection{Accuracy versus Homophily}
For a better understanding of the success of our method, we analyze the relationship between classification accuracy with homophily ratios of nodes. 
On Chameleon dataset, we split the range $[0, 1]$ of homophily ratio into ten segments, and  analyze the percentage of nodes falling in each one. 
What's more, we calculate the classification accuracy for each sub-range.
As shown in Figure \ref{fig:acc}, GCN performs poorly on nodes with low homophily. 
And IDGL's graph structure learner may result in negative effect due to its implicit assumption of homophily. 
In contrast, LGS improves the accuracy of  nodes with strong heterophily without harming performance on nodes with high homophily.

\section{Conclusion}
 In this paper, we introduce a novel label-informed graph structure learning framework (LGS). 
Apart from feature information, LGS incorporates label information into graph structure learning explicitly through a class transition matrix.
We conduct extensive experiments on both homophily and heterophily graph datasets.
Experimental results show that LGS improves the accuracy of nodes with strong heterophily without harming the performance on nodes with high homophily, reflecting the superiority of LGS.


\newpage
\bibliographystyle{ACM-Reference-Format}
\bibliography{ref}

\end{document}